\documentclass{article}

\PassOptionsToPackage{numbers, compress}{natbib}
\usepackage[final]{nips_2018}

\usepackage[utf8]{inputenc} % allow utf-8 input
\usepackage[T1]{fontenc}    % use 8-bit T1 fonts
\usepackage{hyperref}       % hyperlinks
\usepackage{url}            % simple URL typesetting
\usepackage{booktabs}       % professional-quality tables
\usepackage{amsfonts}       % blackboard math symbols
\usepackage{nicefrac}       % compact symbols for 1/2, etc.
\usepackage{microtype}      % microtypography

\usepackage{amsmath,amssymb,amsthm,amsbsy}
\usepackage{graphicx}
\usepackage{subcaption}
\usepackage{multirow}
\usepackage{algorithm}
\usepackage{algorithmic}
\usepackage{enumitem}
\usepackage{wrapfig}

\theoremstyle{remark}
\newtheorem{remark}{Remark}

\newcommand{\integer}{\mathbb{Z}}

\newcommand{\round}{\operatorname{round}}

\title{Universal Deep Neural Network Compression}

\author{
  Yoojin Choi\qquad~Mostafa El-Khamy\qquad~Jungwon Lee\\
  SoC R\&D, Samsung Semiconductor Inc.\\
  San Diego, CA 92121, USA\\
  \texttt{\{yoojin.c,mostafa.e,jungwon2.lee\}@samsung.com} \\
}

\begin{document}
% \nipsfinalcopy is no longer used

\maketitle

\begin{abstract}
In this paper, we investigate lossy compression of deep neural networks (DNNs) by weight quantization and lossless source coding for memory-efficient deployment. Whereas the previous work addressed non-universal scalar quantization and entropy coding of DNN weights, we for the first time introduce universal DNN compression by universal vector quantization and universal source coding. In particular, we examine universal randomized lattice quantization of DNNs, which randomizes DNN weights by uniform random dithering before lattice quantization and can perform near-optimally on any source without relying on knowledge of its probability distribution. Moreover, we present a method of fine-tuning vector quantized DNNs to recover the performance loss after quantization. Our experimental results show that the proposed universal DNN compression scheme compresses the 32-layer ResNet (trained on CIFAR-10) and the AlexNet (trained on ImageNet) with compression ratios of $47.1$ and $42.5$, respectively.%, and with minimal accuracy loss.
\end{abstract}

\section{Introduction} \label{sec:intro}

Compression of deep neural networks (DNNs) has been actively studied in deep learning to develop compact DNN models for memory-efficient and computation-efficient deployment. Han et al.~\cite{han2015deep} showed impressive compression results by weight pruning, $k$-means clustering, and Huffman coding. It is further optimized in \cite{choi2016towards} using Hessian-weighted $k$-means clustering. Recently, it is shown how soft weight sharing or soft quantization can be employed for DNN weight quantization in \cite{ullrich2017soft,agustsson2017soft}. On the other hand, weight pruning is also extensively studied, e.g., in \cite{guo2016dynamic,molchanov2017variational,louizos2017bayesian,lin2017runtime,dai2018compressing}. In this paper, we focus on DNN weight quantization, which can be used together with weight pruning to generate compressed models. 

Vector quantization reduces the gap to the rate-distortion bound by jointly quantizing multiple symbols. Since conjectured by Gersho in \cite{gersho1979asymptotically}, lattice quantization has been presumed to be the most efficient entropy coded vector quantization in the high resolution regime asymptotically, as the rate goes to infinity and the distortion diminishes~\cite{gray1998quantization}. Although lattice quantizers are simple and empirically shown to perform well even at finite rates, their efficiency depends on source statistics. Thus, we consider universal quantization that provides near-optimal performance for any source distribution~\cite{ziv1985universal}. Of particular interest is randomized lattice quantization, where uniform random dithering makes the distortion independent of the source, and the gap of its rate from the rate-distortion bound at any distortion level is provably no more than $0.754$ bits per sample for any finite dimension%, and no more than $0.5$ bits per sample asymptotically as the dimension grows for any sources
~\cite{zamir1992universal}.

From the classical lossy compression results, this paper establishes a universal DNN compression framework consisting of universal quantization and universal lossless source coding such as Lempel--Ziv--Welch~\cite{ziv1977universal,ziv1978compression,welch1984technique} and the Burrows--Wheeler transform~\cite{effros2002universal,seward1998bzip2}. In order to recover any accuracy loss resulting from weight quantization, we furthermore propose a fine-tuning algorithm for vector quantized DNNs. The gain of fine-tuning becomes larger as the vector dimension increases, due to the fact that the number of shared quantized values that are tunable (trainable) in a vector quantized model increases as the vector dimension increases.

\section{Universal DNN compression} \label{sec:unq}

For vector quantization, given $N$ weights denoted by $w_1,\dots,w_N$, we generate $n$-dimensional vectors~$\mathbf{v}_1,\dots,\mathbf{v}_{\lceil N/n\rceil}$ by concatenating $n$ distinct weights into one vector, e.g., as follows:
\begin{equation} \label{sec:unq:eq:01}
\mathbf{v}_i=[w_{ni-n+1}\ w_{ni-n+2}\ \cdots\ w_{ni}]^T,
\ \ \
1\leq i\leq\lceil N/n\rceil,
\end{equation}
where $w_j=0$ for $N+1\leq j\leq\lceil N/n\rceil n$, and $\lceil x\rceil$ is the smallest integer larger than or equal to $x$. Vector quantization partitions these $n$-dimensional vectors into a finite number of clusters, and the vectors in each cluster share their quantized value, i.e., the cluster center.

\begin{figure}[t!]
\centering
\includegraphics[width=0.7\columnwidth]{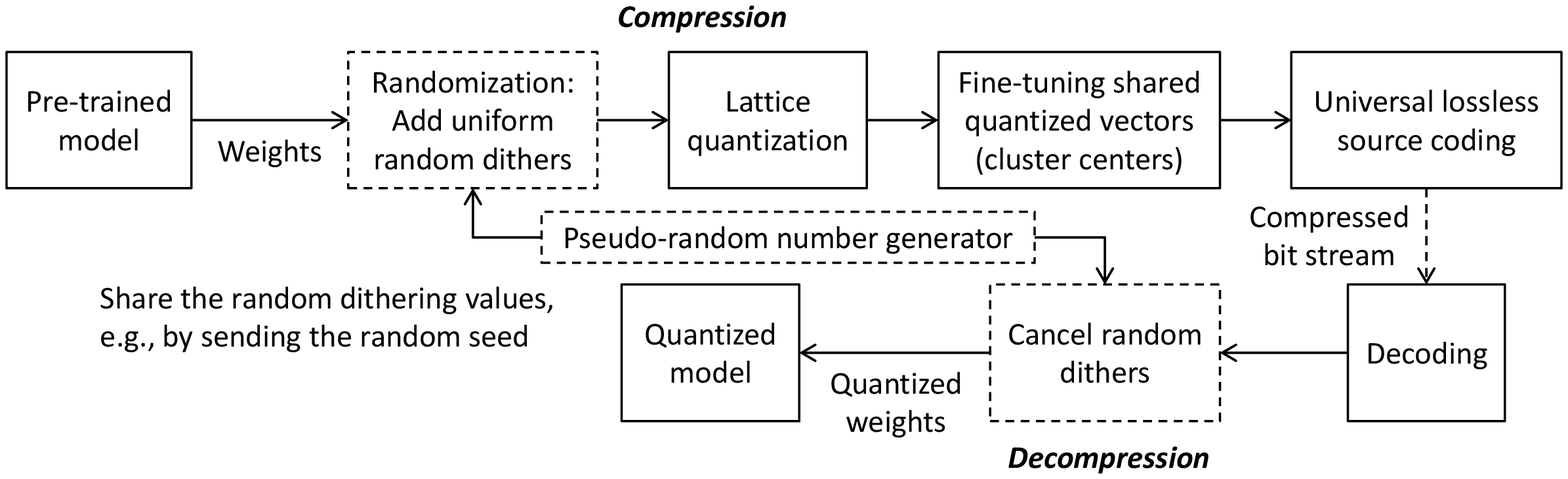}
\caption{Universal DNN compression by universal quantization and universal source coding.\label{sec:unq:random:fig:01}}
\vspace{-1em}
\end{figure}

\textbf{Randomized lattice quantization}: Randomized lattice quantization~\cite{ziv1985universal} achieves universally good performance regardless of source statistics at any rates, and this leads us to the following universal DNN quantization method.

\begin{itemize}[noitemsep,topsep=0em,leftmargin=1.2em]
%\item We first construct $n$-dimensional vectors from weights of all layers by \eqref{sec:unq:eq:01}.
\item We randomize the $n$-dimensional vectors in \eqref{sec:unq:eq:01} by adding uniform random dithers as follows:
%Let $\mathbf{u}_i$ for $1\leq i\leq\lceil N/n\rceil$ be the vectors consisting of random dithers. The randomized $n$-dimensional vectors satisfy
\begin{equation} \label{sec:unq:random:eq:01}
\tilde{\mathbf{v}}_i=\mathbf{v}_i+\mathbf{u}_i,
\ \ \
\mathbf{u}_i=[U_i\ U_i\ \cdots\ U_i]^T,
\ \ \
1\leq i\leq\lceil N/n\rceil,
\end{equation}
where each dithering vector~$\mathbf{u}_i$ consists of $n$ repetitions of a single uniform random variable~$U_i$, and $U_1,\dots U_{\lceil N/n\rceil}$ are independent and identically distributed (i.i.d.) uniform random variables of support $[-\Delta/2,\Delta/2]$; $\Delta$ is the quantization bin size. In each dimension, dithering values are i.i.d. uniform, which is sufficient to make the quantization error independent of source statistics. %The support of the uniform distribution is $[-\Delta/2,\Delta/2]$, where $\Delta$ is the quantization bin size.
\item After dithering, uniform quantization in each dimension (i.e., lattice quantization) follows, i.e.,
\begin{equation} \label{sec:unq:random:eq:02}
\mathbf{q}_i=\Delta\round(\tilde{\mathbf{v}}_i/\Delta),
\end{equation}
where $\mathbf{q}_i$ is the quantized vector of $\tilde{\mathbf{v}}_i$; the rounding and the scaling are element-wise operations in \eqref{sec:unq:random:eq:02}, where the rounding yields the closest integer values of the input.
\end{itemize}

\begin{remark} \label{sec:unq:remark:01}
Vector quantization theoretically provides a better rate-distortion trade-off. However, in practice, for compression of a finite number of data, the gain of vector quantization is limited by the codebook overhead, which becomes more considerable as dimension~$n$ increases and becomes the dominant factor that degrades the compression ratio after some point (see Figure~\ref{sec:exp:fig:01} and compare the cases of $n=4$ and $n=6$).
\end{remark}

\textbf{Fine-tuning of vector quantized DNNs}: We fine-tune the vector quantization codebook to recover the loss after quantization. Each element of a shared quantized vector in the codebook is fine-tuned separately. That is, if we have $k_\textit{VQ}$ clusters of $n$-dimensional vectors, we effectively divide weights into $nk_\textit{VQ}$ groups and fine-tune their shared quantized values separately. The average gradient of the network loss function with respect to weights is computed in each group and used to update their shared quantized value, as will be clarified below.

Let $\mathbf{c}_i=[c_{i,1}\ \cdots\ c_{i,n}]^T$ be the shared quantized vector present in the codebook for cluster~$i$, and let $\mathcal{I}_{i,j}$ be the index set of all weights that are quantized to the same value~$c_{i,j}$ from \eqref{sec:unq:random:eq:02}, for $1\leq i\leq k_\textit{VQ}$ and $1\leq j\leq n$. The gradient descent for the shared quantized value~$c_{i,j}$ is then given by
\begin{equation}
c_{i,j}(t)
=c_{i,j}(t-1)-\eta\frac{1}{|\mathcal{I}_{i,j}|}\sum_{l\in\mathcal{I}_{i,j}}\nabla_{w_l}L(t-1),
\end{equation}
where $t$ is the iteration time, $L$ is the network loss function, and $\eta$ is the learning rate. The individual quantized vectors from \eqref{sec:unq:random:eq:02} are also updated following their shared quantized vectors in the codebook.

In randomized lattice quantization, we would like to clarify that the average gradients are computed from the network loss function evaluated at the quantized weights after canceling random dithers (see \eqref{sec:unq:random:eq:03}), while we fine-tune the shared values obtained before canceling random dithers (see \eqref{sec:unq:random:eq:02}).

\textbf{Universal source coding}: Universal source coding algorithms are more convenient in practice than Huffman coding since they do not require us to estimate source statistics as in Huffman coding. Moreover, they utilize dictionary-based coding, where the codebook (i.e., dictionary) is built from source symbols adaptively in encoding and decoding, and therefore the codebook overhead is smaller than Huffman coding.% (see Remark~\ref{sec:unq:remark:01}).

For universal compression, the indices in the codebook of the lattice quantization output are passed as an input stream to a universal source coding scheme, which produces a compressed stream. The decoder needs to deploy the codebook that contains the indices and their corresponding fine-tuned shared quantized values for decompression.

\textbf{Decompression}: Using randomized lattice quantization, the encoder and the decoder are assumed to share the information on random dithers or their random seed. Under this assumption, the decoder decompresses the fine-tuned lattice quantization output from the compressed stream and then cancels the dithering vectors to obtain
\begin{equation} \label{sec:unq:random:eq:03}
\bar{\mathbf{v}}_i=\mathbf{q}_i-\mathbf{u}_i,
\end{equation}
which yields the deployed weights of the universally quantized DNN at the inference step.

\section{Experiments and discussion} \label{sec:exp}

\begin{figure}
\centering
{\scriptsize
\begin{minipage}[b]{0.45\columnwidth}
  \centering
  \centerline{\includegraphics[width=\columnwidth]{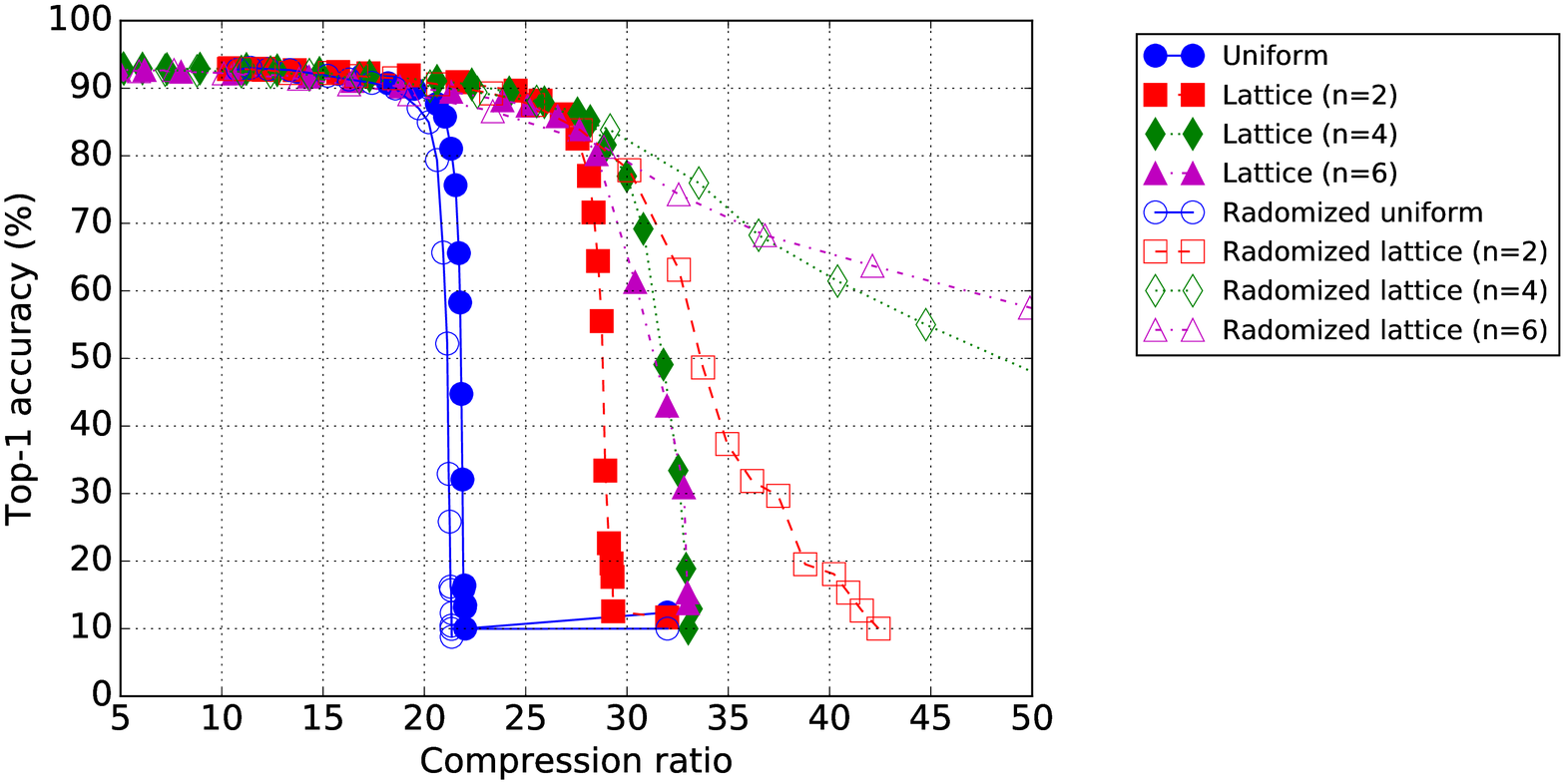}}
  \centerline{(a) Uniform boundaries $\{i\Delta,i\in\integer\}$}
\end{minipage}
\quad
\begin{minipage}[b]{0.45\columnwidth}
  \centering
  \centerline{\includegraphics[width=\columnwidth]{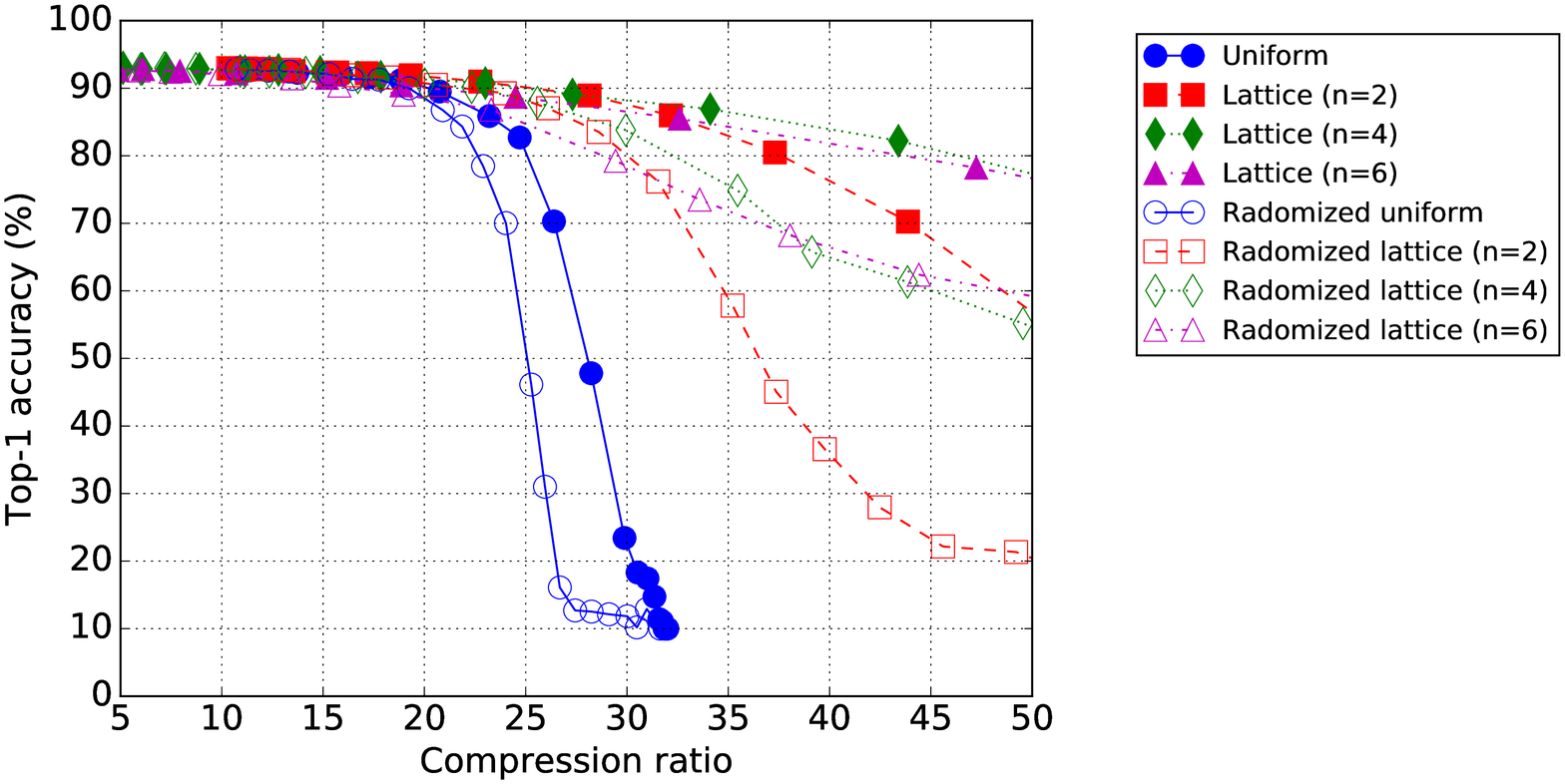}}
  \centerline{(b) Uniform boundaries $\{(2i+1)\Delta/2,i\in\integer\}$}
\end{minipage}
}
\caption{Compression results of 32-layer ResNet by (randomized) lattice quantization and Huffman coding for different values of dimension~$n$. No weight pruning is used in the results of this figure.\label{sec:exp:fig:01}}
%, where the uniform boundaries in each dimension are set from (a) $\{i\Delta,i\in\integer\}$ and (b) $\{(2i+1)\Delta/2,i\in\integer\}$, respectively.\label{sec:exp:fig:01}}
\vspace{-1em}
\end{figure}

\setlength{\tabcolsep}{0em}
\begin{wraptable}{r}{21em}
\vspace{-4.5em}
\caption{Compression of pruned models.\label{sec:exp:res:pruned:tlb:01}}
\centering
{\scriptsize
\begin{tabular}{llcc}
Model & Method & \shortstack[c]{Top-1\\accuracy (\%)} & \shortstack[c]{Compression\\ratio} \\
\toprule
\multirow{9}{*}{\shortstack[l]{ResNet-32\\(CIFAR-10)\ \ \ }}
& Pre-trained model                                    & 92.58 &     - \\
\cmidrule{2-4}
& Ours: Weight pruning (87.5\%)*                       & 92.67 & 7.07  \\
& \ \ \ + Universal quantization + bzip2               & 92.57 & 32.87 \\
\cmidrule{2-4}
& Ours: Weight pruning (92.7\%)*                       & 92.06 & 11.86 \\
& \ \ \ + Universal quantization + bzip2               & 92.06 & 47.10 \\
\cmidrule{2-4}
& Han et al.~\cite{han2015deep}                        &   N/A &   N/A \\
& Choi et al.~\cite{choi2016towards}                   & 92.68 & 22.17 \\
& Agustsson et al.~\cite{agustsson2017soft}            & 92.10 & 20.15 \\
\midrule                                                                                            
\multirow{7}{*}{\shortstack[l]{AlexNet\\(ImageNet)}}
& Pre-trained model                                    & 57.22 &     - \\
\cmidrule{2-4}
& Ours: Weight pruning (90.4\%)*                       & 57.28 & 9.11  \\
& \ \ \ + Universal quantization + bzip2               & 57.02 & 42.46 \\
\cmidrule{2-4}
& Han et al.~\cite{han2015deep}                        & 57.22 & 35.00 \\
& Choi et al.~\cite{choi2016towards}                   & 56.20 & 40.65 \\
& Agustsson et al.~\cite{agustsson2017soft}            & N/A   & N/A   \\
\bottomrule
\multicolumn{4}{r}{* These rows show the compression results from weight pruning only.}
\end{tabular}
}
\vspace{-1em}
\end{wraptable}

We evaluate the proposed DNN compression scheme first without pruning for the 32-layer ResNet~\cite{he2016deep} (ResNet-32) model %\footnote{https://github.com/Lasagne/Recipes/tree/master/papers/deep\_residual\_learning}
on CIFAR-10 dataset~\cite{krizhevsky2009learning}. We consider two cases for (randomized) lattice quantization where the uniform boundaries in each dimension are set from (a) $\{i\Delta,i\in\integer\}$ and (b) $\{(2i+1)\Delta/2,i\in\integer\}$, respectively; $\integer$ is the set of integers. The quantization bin size~$\Delta$ is the same for both cases, but case~(a) has the zero at a bin boundary while in case~(b) the zero element is at the center of the middle bin.%, which removes the bias.

For unpruned models, we often have a high volume of weights concentrated around zero, and thus case~(b) that assigns one bin to include all the weights near zero is expected to outperform case (a), which is aligned with our lattice quantization results in Figure~\ref{sec:exp:fig:01}. However, it is interesting to observe that randomized lattice quantization provides similarly good performance in both cases, which is the main benefit of randomizing the source by uniform dithering before quantization. %We emphasize that randomized lattice quantization is applicable for any network models with no optimization regardless of the statistics of their weights.
Figure~\ref{sec:exp:fig:01} also shows that vector quantization provides additional gain over scalar quantization particularly when the compression ratio is large.

Finally, Table~\ref{sec:exp:res:pruned:tlb:01} summarizes the compression ratios that we obtain from our universal DNN compression method for pruned ResNet-32 and AlexNet~\cite{krizhevsky2012imagenet} models. The proposed universal DNN compression scheme with the \emph{bzip2}~\cite{seward1998bzip2} universal source coding algorithm yields $47.10\times$ and $42.46\times$ compression for ResNet-32 and AlexNet, respectively. %, which are better than the previous work~\cite{han2015deep,choi2016towards,agustsson2017soft}.
Compared with \cite{han2015deep,choi2016towards,agustsson2017soft} which need to optimize and/or calculate source statistics for compression, we achieved a better trade-off between rate (compression ratio) and distortion (loss in accuracy) through the universal compression of DNNs.

{\small
\newcommand{\BIBdecl}{\setlength{\itemsep}{0.25em}}
\bibliographystyle{IEEEtran}
\bibliography{manuscript}
}

\newpage
\appendix

\section{Appendix}

\begin{figure}[t]
\centering
{\scriptsize
\begin{minipage}[b]{0.45\columnwidth}
  \centering
  \centerline{\includegraphics[width=\columnwidth]{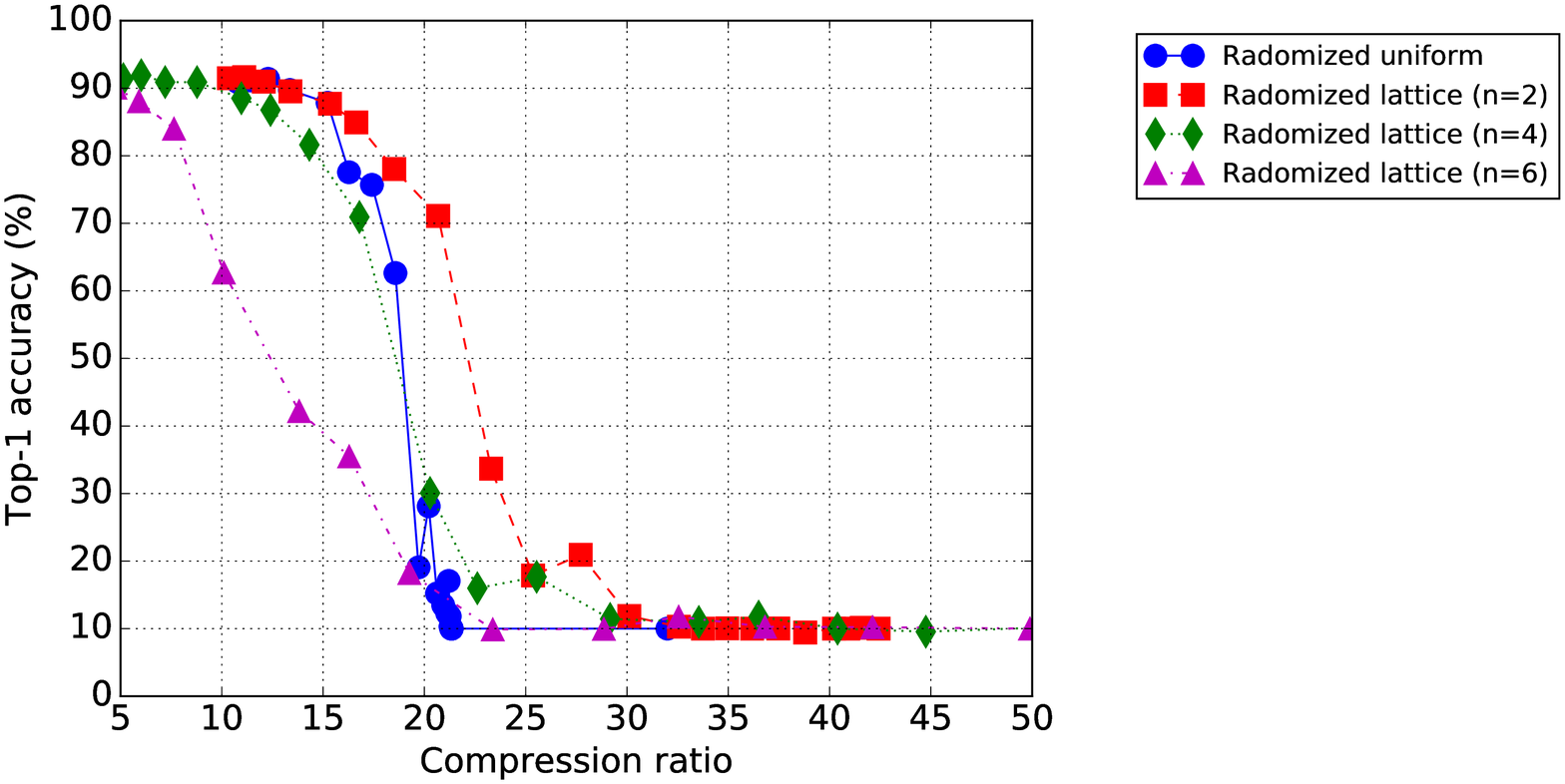}}
  \centerline{(a) Before fine-tuning the quantization codebook}
\end{minipage}
\quad
\begin{minipage}[b]{0.45\columnwidth}
  \centering
  \centerline{\includegraphics[width=\columnwidth]{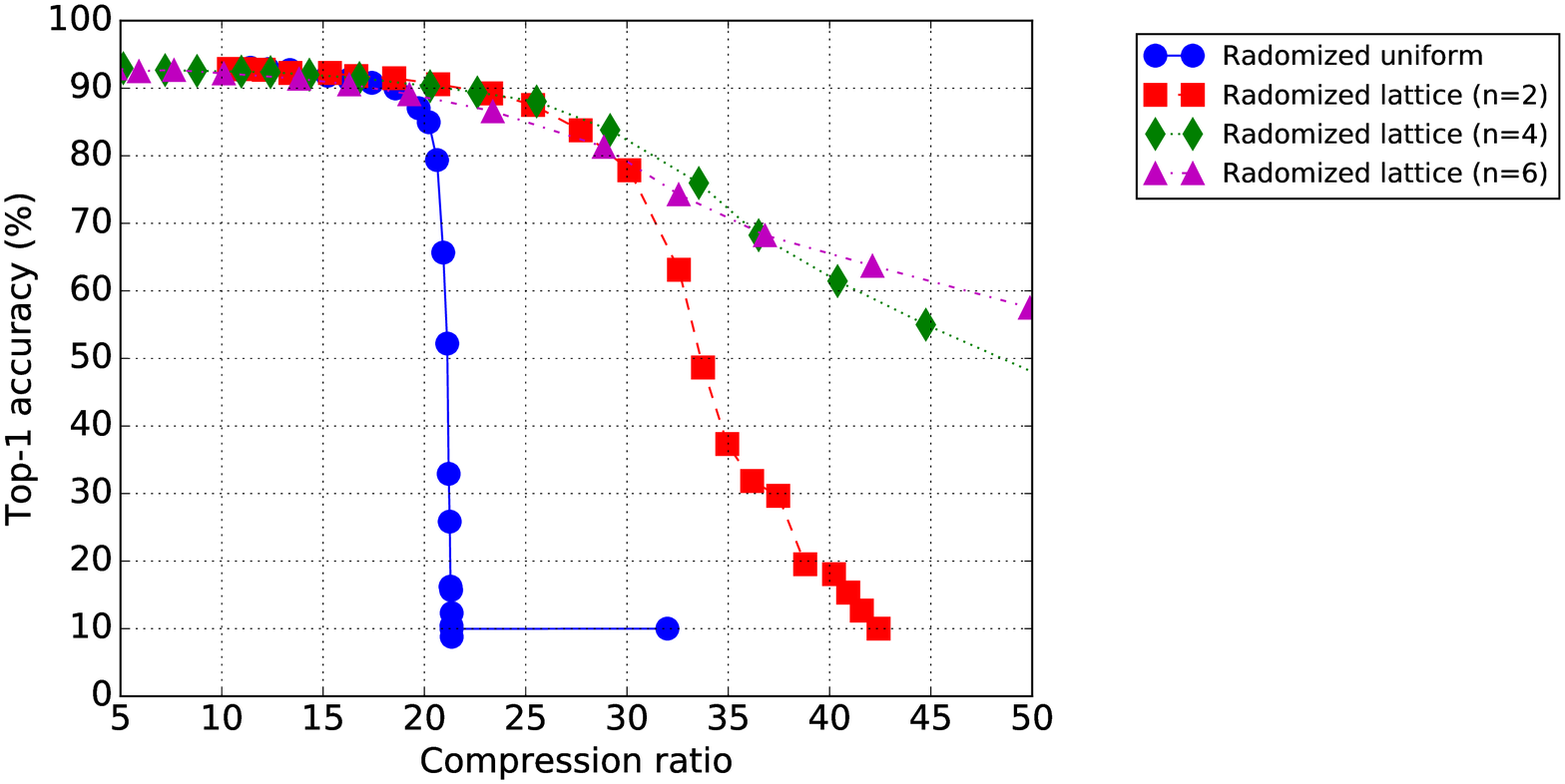}}
  \centerline{(b) After fine-tuning the quantization codebook}
\end{minipage}
}
\caption{Performance comparison before and after fine-tuning the codebook for the 32-layer ResNet models, quantized with randomized lattice quantization for different values of vector dimension~$n$, where the uniform boundaries in each dimension are from $\{i\Delta,i\in\integer\}$. \label{sec:exp:res:vq:resnet32:fig:04}}
\vspace{-1em}
\end{figure}
\setlength{\tabcolsep}{.5em}
\begin{table}[t]
\caption{Additional compression results for the pruned AlexNet model.\label{sec:exp:res:pruned:tlb:02}} \vspace{.5em}
\centering
{\scriptsize
\begin{tabular}{llcc}
Model & Method & Top-1 accuracy (\%) & Compression ratio \\
\toprule                                          
\multirow{5}{*}{\shortstack[l]{AlexNet\\(ImageNet)}}
& Pre-trained model                                                      & 57.22 &     - \\
%\cmidrule{2-4}
& Weight pruning (90.4\%)                                                & 57.28 &  9.11 \\
& Weight pruning (90.4\%) + Universal quantization + Huffman coding      & 57.02 & 41.43 \\
& Weight pruning (90.4\%) + Universal quantization + LZW                 & 57.02 & 41.82 \\
& Weight pruning (90.4\%) + Universal quantization + bzip2               & 57.02 & 42.46 \\
\bottomrule
\end{tabular}
}
\vspace{-1em}
\end{table}

\subsection{Experimental details for the results in Figure~\ref{sec:exp:fig:01}}

Figure~\ref{sec:exp:fig:01} shows the compression results of the unpruned ResNet-32 model. In particular, we compare the trade-off between the accuracy and the compression ratio for lattice quantization and randomized lattice quantization with different values of vector dimension~$n$ (see \eqref{sec:unq:eq:01}); here, uniform quantization corresponds to lattice quantization with dimension~$n=1$. Given the vector dimension~$n$, the weights from all layers of the pre-trained ResNet-32 model are vectorized as in \eqref{sec:unq:eq:01} for vector quantization. Then, lattice quantization or randomized lattice quantization follows. In plain lattice quantization, no random dithering is added before quantization, i.e., we set $\mathbf{u}_i=\mathbf{0}$ for all $i$ in \eqref{sec:unq:random:eq:01}. We fine-tune the quantization codebook as explained in Section~\ref{sec:unq}. We simply use Huffman coding only in this experiment to get the compressed models. %The compression ratio is the ratio of the total number of bits for the original weights to the number of bits for the compressed ones and the codebook, when the original weights are 32-bit floating-point numbers.

The gain of randomized lattice quantization over lattice quantization can be found in Figure~\ref{sec:exp:fig:01}(a) in particular for $n\geq2$ and large compression ratios. We note that randomized lattice quantizers provide similarly good performance in both cases (a) and (b). Lattice quantization performs well only in case (b), where the quantization bins are optimized for given weight distribution. We emphasize that randomized lattice quantization is applicable for any network models blindly, regardless of their weight distribution and with no optimization, while it is guaranteed to yield a good rate-distortion trade-off close to the optimum within a fixed gap~\cite{ziv1985universal}.

\subsection{Additional results that show the impact of fine-tuning}

In Figure~\ref{sec:exp:res:vq:resnet32:fig:04}, we show the performance of universally quantized ResNet-32 models before and after fine-tuning the codebook. The gain of vector quantization becomes more significant after fine-tuning, in particular, as the vector dimension~$n$ increases, since there are a more number of shared quantized values trainable in vector quantized models.

\subsection{Contribution of pruning, quantization, and source coding}

Table~\ref{sec:exp:res:pruned:tlb:02} shows the incremental improvement of the compression ratio for the AlexNet model. Given a pre-trained AlexNet model, we prune its $90.4$\% weights and fine-tune remaining unpruned weights, as suggested in \cite{han2015learning}, which yields the compression ratio of $9.11$ with the top-1 accuracy of $57.28$\%. %The accuracy improves slightly after fine-tuning the pruned model.
Using universal quantization and \emph{bzip2} universal source coding for the pruned AlexNet model, we achieve the compression ratio of $42.46$ with the top-1 accuracy of $57.02$\%. %We fine-tune the codebook and recover the top-1 accuracy of $57.02$\% from $56.80$\% before fine-tuning.
We also compare Huffman coding to the universal source coding algorithms, i.e., Lempel--Ziv--Welch (LZW) and \emph{bzip2} that deploys the Burrows--Wheeler transform~\cite{effros2002universal}. Both LZW and \emph{bzip2} provide better compression ratios than Huffman coding in our experiments.

\end{document}